%% file: main.tex
\documentclass[10pt,twocolumn,letterpaper]{article}

\usepackage{wacv}
\usepackage{times}
\usepackage{epsfig}
\usepackage{graphicx}
\usepackage{amsmath}
\usepackage{amssymb}
\usepackage{url}
\usepackage{paralist}
\usepackage{setspace} 
\usepackage{multirow}
\usepackage{subcaption}
\usepackage{booktabs}
\usepackage{algorithm}
\usepackage[noend]{algpseudocode}

\DeclareMathOperator*{\argmax}{arg\,max}

\usepackage{IEEEtrantools}

\wacvfinalcopy 

\def\wacvPaperID{325} 

\ifwacvfinal\fi
\begin{document}

\title{Prototypical Clustering Networks for Dermatological Disease Diagnosis}


\author{Viraj Prabhu\thanks{Work done while V.P. was an intern at Curai.}$\,\,^{,1}$
\qquad
Anitha Kannan$^{3}$ \qquad Murali Ravuri$^{3}$ \\ Manish Chablani$^{3}$ \qquad David Sontag\thanks{Work done as an advisor to Curai.}$^{2}$ \qquad Xavier Amatriain$^{3}$ \qquad \\
$^1$Georgia Tech \qquad $^2$MIT \qquad $^3$Curai \\
{\tt\small{virajp@gatech.edu} \qquad\tt\small{dsontag@csail.mit.edu} \qquad \tt\small \{anitha, murali, manish, xavier\}@curai.com}
}

\maketitle
\ifwacvfinal\thispagestyle{empty}\fi


\begin{abstract}
We consider the problem of image classification for the purpose of aiding doctors in dermatological diagnosis. Dermatological diagnosis poses two major challenges for standard off-the-shelf techniques: First, the data distribution is typically extremely long tailed. Second, intra-class variability is often large. To address the first issue, we formulate the problem as low-shot learning, where once deployed, a base classifier must rapidly generalize to diagnose novel conditions given very few labeled examples. To model diverse classes effectively, we propose Prototypical Clustering Networks (PCN), an extension to Prototypical Networks~\cite{snell2017prototypical} that learns a mixture of prototypes for each class. Prototypes are initialized for each class via clustering and refined via an online update scheme. Classification is performed by measuring similarity to a weighted combination of prototypes within a class, where the weights are the inferred cluster responsibilities. We demonstrate the strengths of our approach in effective diagnosis on a realistic dataset of dermatological conditions.
\end{abstract}


\input{source/intro}
\input{source/relwork}
\input{source/approach}

\input{source/experiments}

\input{source/conclusion}

{\small
\bibliographystyle{ieee}
\bibliography{wacv}
}
\input{source/appendix}

\end{document}

%% file: source/intro.tex
\vspace{-20pt}

\section{Introduction}
Globally, skin disease is one of the most common human illnesses that affects 30\% to 70\% of individuals, with even higher rates in at-risk subpopulations where access to care is scarce \cite{NHANES78, Bickers06, Scholfield09, Hay12, Basra09}. Untreated or mistreated skin conditions often lead to detrimental effects including physical disability and death \cite{Basra09}.

A large fraction of skin conditions are diagnosed and treated at the first point of contact, {\it i.e.} by primary care and general practitioners. While this makes access to care faster, recent studies indicate that general physicians, especially those with limited experience, may not be well-trained for diagnosing many skin conditions \cite{Federman99, Goldman07}. In addition, people with no or little access to health care systems often depend on their own search and `image recognition capabilities' to self (mis-)diagnose and treat. While there is a recent surge in online services and telemedicine for closing the gap of healthcare access, these services also have similar problems~\cite{Resneck16}. 
The need to find effective solutions to \emph{aid} doctors in accurate diagnosis motivates this work. 

Why is diagnosis of skin conditions hard for doctors? One important factor is the sheer number of dermatological conditions. The International Classification of Disease 10 (ICD 10) classification of  human disease\footnote{http://www.who.int/classifications/icd/en/} enumerates more than 1000 skin or skin-related illnesses. 
However, most general physicians are trained on a few tens of common skin ailments under the assumption that this will enable accurate diagnoses in most cases. Recent studies indicate that this assumption may be flawed~\cite{Wilmer14}. To make an accurate diagnosis, the knowledge of all possible diseases becomes important, especially to workup and eliminate possible life-threatening conditions. The difficulty of diagnosis is further compounded by the large intra-class variability within several conditions. To motivate the scale of this problem, see Figure~\ref{fig:teaser}, where we show the class distribution of Dermnet\footnote{http://www.dermnet.com/}, a publicly available large-scale dataset of dermatological conditions. 
The plot shows examples illustrating the intra-class variability found in the dataset. This makes accurate diagnosis challenging even for experienced dermatologists.

These issues create an opportunity for incorporating automated machine learning systems as part of the doctor's workflow, aiding them in sieving through possible skin conditions. AI systems have shown promising results in many applications in computer vision (see {\emph{c.f.} \cite{he2016deep},\cite{huang2017densely} and citations within). These advances have started to impact the healthcare domain, with early applications on automated classification of skin lesions using images \cite{esteva2017dermatologist} and diagnosis based on radiology data \cite{Li17}.

Inspired by these recent successes, this paper tackles the problem of fine-grained skin disease classification. We conjecture that a high fidelity AI system can serve as a diagnostic decision support system to general physicians. By suggesting candidate diagnoses, it can greatly reduce effort and compensate for the possible lack of experience or time at the point of care. In the context of teledermatology with a store-and-forward approach that involves asynchronous evaluation by dermatologists, such a system can aid in triaging the right doctor resource in a timely manner, especially when acute conditions need immediate care \cite{Goldman07}.

Learning a model for dermatological image classification poses some major challenges:
\begin{itemize}
\item \textbf{Access.} Access to large amounts of data may not always be possible. As dermatology images are collected as part of Electronic Health Records (EHR), access is usually strictly controlled for privacy reasons. For a new healthcare platform that wants to build a dermatological classifier, starting with a small set of  conditions and rapidly increasing the scope of predictable diagnoses is often the only practical alternative.

\item \textbf{Long tail.} The data distribution is invariably long tailed. Some skin conditions are rare and may not have many recorded examples. Others may be common but are so easily diagnosable that they are simply not recorded in EHR. In Figure~\ref{fig:teaser}, notice how common conditions such as flea bites and rarer diseases such as melanoma both end up in the tail of the dataset. 
\item \textbf{Intra-class diversity.} Several conditions contain significant intra-class variability, for eg. a condition like acne may occur on the face, back, scalp, etc. 
\end{itemize}

Despite these issues, we need robust mechanisms to make correct diagnoses. Our approach pursues the following objectives: First, the model needs to be able to handle the long-tail in the data and perform well on classes in both the head and the tail. Second, once deployed, it needs to be easily extensible to novel classes that it encounters given very few labeled examples (potentially labeled by a physician). With these objectives, we pose dermatological image classification as a few-shot learning problem. Our proposed model, that we call Prototypical Clustering Networks (PCN), extends prior work on Prototypical Networks~\cite{snell2017prototypical} to represent a class as a mixture of prototypes instead of a single prototype. Training this classifier involves learning an embedding space while simultaneously learning to represent each class as a mixture of prototypes. Prototypes are initialized for each class via clustering and refined via an online update scheme. Classification is performed by measuring similarity to a weighted combination of prototypes within a class, where the weights are the inferred cluster responsibilities.
The examples shown in Figure~\ref{fig:teaser} are in fact, nearest neighbors to prototypes of the classes learned using the proposed approach. We extensively compare the performance of the algorithm to Prototypical Networks and other strong baselines on Dermnet.

\begin{figure}[t]
 \centering 
 \includegraphics[width=1\linewidth]{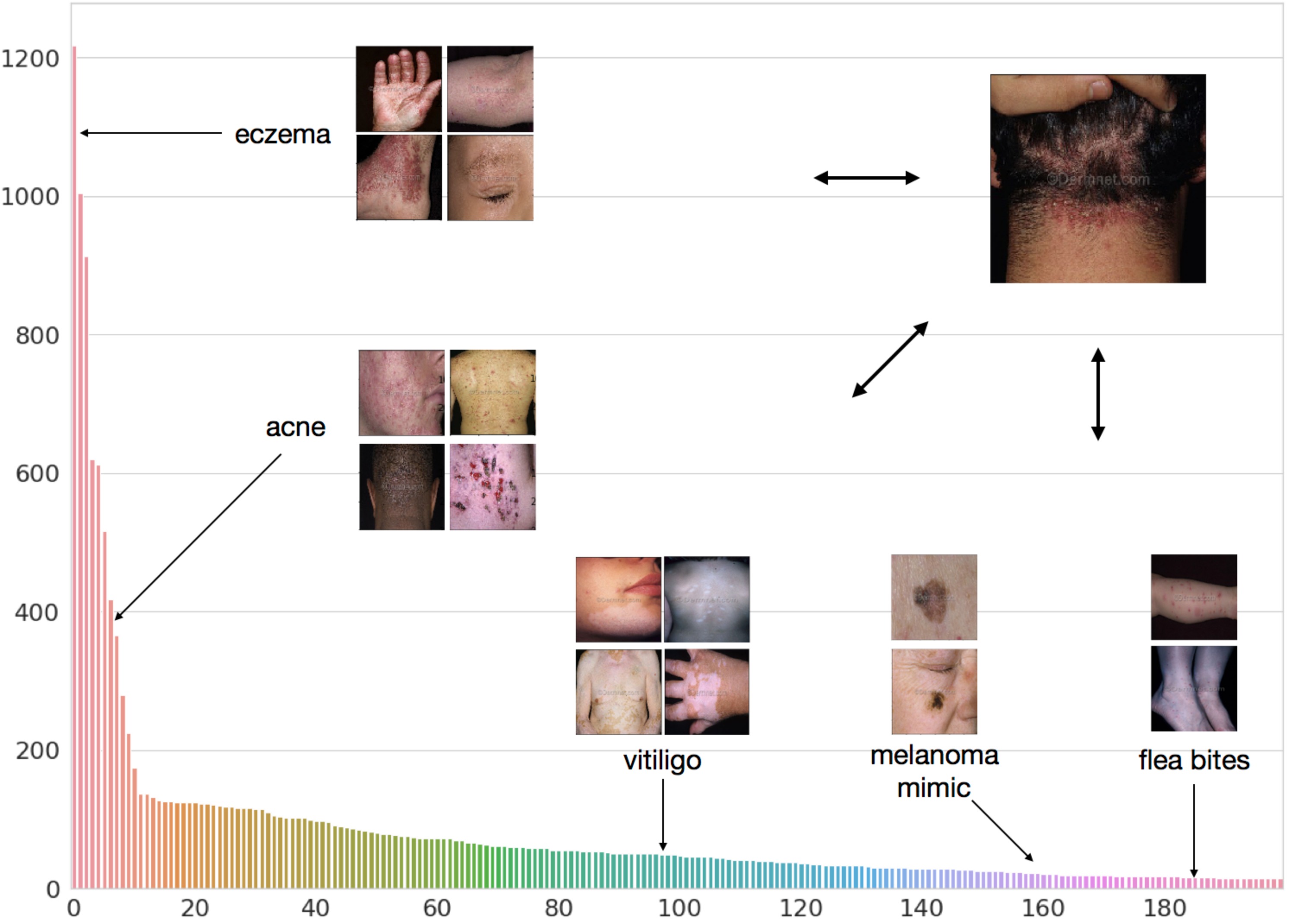}
 \vspace{-18pt}
 \caption{
Long-tailed class distribution of Dermnet (shown here for the top-200 classes). Also shown are nearest neighbors to four of the many prototypes learned for select classes using the proposed Prototypical Clustering Network approach. This is illustrative of the huge intra-class variability in the data. For a novel test image, shown at the upper right corner, the model predicts the correct class by measuring weighted similarity to per-class clusters in the embedding space learned through a deep convolutional neural network.
 }
\vspace{-13pt}
\label{fig:teaser}
\end{figure}

%% file: source/relwork.tex
\section{Related Work}

\par \noindent
\textbf{Dermatological Classification.} A few prior works address the problem of dermatological classification. In~\cite{esteva2017dermatologist}, authors focus specifically on diagnosing skin cancer, and establish a benchmark on a large closed-source dataset of skin lesions by finetuning a pretrained deep convolutional neural network (CNN). In~\cite{liao2016deep}, authors study the problem of skin disease diagnosis on the Dermnet dataset but focus on coarse 23-way classification using its top-level hierarchy. In this work, we study fine-grained recognition of skin conditions on the Dermnet dataset in a few-shot setup, and propose a method to model multimodal classes and generalize effectively to unseen novel classes with very little data. 
\par \noindent
\textbf{Class-imbalanced datasets.} Real world datasets typically possess long tails~\cite{van2017devil,wang2017learning,zhu2014capturing}, and learning robust CNN representations from such data is a topic of active research. Conventional training methods typically lead to poor generalization on tail classes as class-prior statistics are skewed towards the head of the distribution. Simple techniques such as random oversampling (or undersampling) by repeating (or removing) tail instances are found to help mitigate this issue to a degree~\cite{buda2018systematic}. Alternative approaches include \cite{wang2017learning}, that proposes a meta learning algorithm to transfer knowledge from data-rich head classes to the tail. In this work, we propose a few-shot learning approach on a real-world imbalanced dataset of dermatological conditions, and demonstrate strong generalization capabilities even in the presence of very few training examples.
\par \noindent
\textbf{Few-shot learning.} Few-shot learning aims to learn good class representations given very few training examples~\cite{snell2017prototypical,vinyals2016matching,triantafillou2018meta,hariharan2017low,santoro2016one}. Main paradigms of approaches include simulating data starved environments at training time, and including non-parametric structures in the model as regularizers. Matching networks\cite{vinyals2016matching} learn an attention mechanism over support set labels to predict query set labels for novel classes. Prototypical networks\cite{snell2017prototypical} jointly learn an embedding and centroid representations (as class \textit{prototypes}), that are used to classify novel examples based on Euclidean distance. In both \cite{vinyals2016matching} and \cite{snell2017prototypical}, embeddings are learned end-to-end and training employs episodic sampling. In an incremental few-shot learning context, \cite{rebuffi2017icarl} propose a class-centroid based representation in an embedding space learned using a generative model. In \cite{hariharan2017low}, authors study few-shot learning on an imbalanced dataset, treating tail classes as novel, and propose a method to ``hallucinate'' additional samples for such data-starved classes. In this work, we focus on a similar setup on the real-world long-tailed Dermnet dataset. We propose an extension to \cite{snell2017prototypical} to model the multimodal nature of diverse classes, and demonstrate how this also helps generalize better to data-starved novel classes. \\
\par \noindent
\textbf{Prototypical Networks.} Prior extensions to Protoypical Networks exist in the literature, and here we distinguish our contributions~\cite{triantafillou2018meta, fort2017gaussian}. In \cite{triantafillou2018meta}, authors propose extending prototypical networks to a semi-supervised setting by using unlabeled examples while producing prototypes. In \cite{fort2017gaussian}, authors propose additionally predicting a covariance estimate for each embedding and using a direction and class dependent distance metric instead of euclidean distance. In this work, we extend prototypical networks to model multimodal classes in an automated diagnostic setting by learning multiple prototypes per class, that are initialized via clustering and refined via an online update scheme.
\par \noindent

%% file: source/approach.tex
\vspace{-15pt}
\section{Approach}
\label{sec:approach}

\begin{algorithm*}
\begin{algorithmic}[1]
\State {\bf Input}:Training set $\mathcal{D} = \{(x_1, y_1), \cdots ,(x_N , y_N )\}$, where each $y_i \in \{1,\cdots, K\}$. $\mathcal{D}_k$ denotes the
subset of $\mathcal{D}$ containing all class prototypes, i.e. elements $(x_i, y_i) =  \{\mu_{z,k}\}_{z=1}^{M_k} \forall k \in  \{1,\cdots, K\}$
\State {\bf Output}: The loss J for a randomly generated training episode  \\
\State V $\leftarrow$ RANDOMSAMPLE($\{1,\cdots, K\}$, $N_C$)
\Comment{Select class indices for episode}
 \For{$k \in \{1,\cdots, N_C\}$} 
 \State $S_k \leftarrow$ RANDOMSAMPLE($\mathcal{D}_{v_k}$, $N_S$) \Comment{Select support examples}
 \State $Q_k \leftarrow$ RANDOMSAMPLE($\mathcal{D}_{v_k \setminus S_k}$, $N_Q$) \Comment{Select query examples}
 {\color{blue}
 \For{$(x, y) \in S_k $} \Comment{Compute probabilistic assignment of x to y's clusters}
 	\State $q(z|k,x)  = \frac{\exp(-d(f_\phi(x), \mu_{z,k})/\tau)}{\sum_{z'} \exp( -d(f_\phi(x), \mu_{z',k})/\tau)}$ 
     	\EndFor }
\EndFor
\For{$z \in \{1,\cdots, M_k \} $}
 {\color{blue}
    \State $\mu_{z,k}^{new}  \leftarrow \alpha \mu_{z,k}^{old} + (1-\alpha) \frac {\sum_{(x,y)\in S_k} q(z | k,x) f_\phi(x)} {\sum_{(x,y)\in S_k} q(z | k,x)}$ }
\EndFor 
\State $L_{\phi} \leftarrow 0 $ 
\For{$k \in \{1,\cdots, N_C\}$}
 {\color{blue}
 	\For{$(x, y) \in Q_k $}
		\State $L_{\phi}$ $\leftarrow$ $L_{\phi}$ + $\frac{1}{N_C N_Q}$ \Big[$\sum_{z} q(z | k,x) d(f_\phi(x), \mu_{z,k})  + \log \sum_{k'} \exp(- \sum_{z'} q(z' | k',x) d(f_\phi(x), \mu_{z',k'}))$ \Big] \label{eqn:loss}
 	\EndFor }
 \EndFor
\caption{Training episode loss computation for Prototypical Clustering Networks. N is the number of examples in the training set, $K_{base}$ is the number of base classes for training, $M_k$ is the number of clusters for class $k$, $N_C \leq K_{base}$  is the number of classes per episode, $N_S$ is the number of support examples per class, $N_Q$ is the number of query examples per class. RANDOMSAMPLE(S, N) denotes a set of N elements chosen uniformly at random from set S, without replacement. {\color{blue}{\it Differences from Algorithm 1 in \cite{snell2017prototypical} in blue}}}
  \label{algo:episodic_training}
 \end{algorithmic}
\end{algorithm*}

We formulate dermatological image classification as a low-shot learning problem. During training time, we have access to a labeled dataset of images $S = \{(\mathbf{x}_1, y_1), ... , (\mathbf{x}_N, y_N)\}$ where each $x_i$ is an observation and $y_i \in \{1, ..., K_{base}\}$ is the label mapping to one of the \emph{base} classes known at training time. At test time,  we are also provided with a small labeled dataset corresponding to $K_{novel}$ novel classes, and must learn to perform $K_{base+novel}$ way classification.  

\subsection{Model}
Prototypical Clustering Networks (PCN) builds upon recent work in Prototypical Networks \cite{snell2017prototypical}. PCN represents each class using a set of prototypical representations learned from the data.  Let $\{\mu_{z,k}\}_{z=1}^{M_k}$ be the collection of $M_k$ prototypes for class $k$. Then, at test time, we  measure similarity to these representations to derive its corresponding class label. In particular, 
\begin{equation}
p(y=k|\mathbf{x}; \phi) = \frac{\exp(-\sum_{z} q(z | k, x) d(f_\phi(x), \mu_{z,k}))} {\sum_{k'} \exp(-\sum_{z'} q(z' | k', x) d(f_\phi(x), \mu_{z',k'}))}
\label{eq:pyGx}
\end{equation}
where $f_\phi(x)$ is the embedding function with learnable parameters $\phi$ that maps input $x$ to a learned representation space, $d$ is a distance function and $q(z | k, x)$  (eq.~\ref{eq:clusterposterior}) is soft assignment of examples to clusters from the class. When $M_k =1$ for all classes, we revert to prototypical networks. 

\subsubsection{Model training}
The goal is to learn a model with parameters $\phi$ so as to maximize the likelihood of the correct class:
\begin{equation}
\phi^{*} = \argmax_{\phi} \sum_{(\mathbf{x}, y)} \log p(y | \mathbf{x}; \phi), 
\end{equation}
and minimize its corresponding loss function $L_{\phi}$. We use  episodic training~\cite{snell2017prototypical,vinyals2016matching,ravi2016optimization} to learn the embedding function by optimizing the loss  and updating the cluster prototypes for each class.  In particular, a training epoch consists of $E$ episodes. Algo.~\ref{algo:episodic_training} provides the details of computing the loss for one episode that is used in learning the function. We describe key components of the algorithm below: \\
{\bf Class-specific cluster responsibilities:}
The assignment of an example within each class is given by:
\begin{equation}
q(z|k,x)  = \frac{\exp(-d(f_\phi(x), \mu_{z,k})/\tau)}{\sum_{z'} \exp(-d(f_\phi(x), \mu_{z',k})/\tau)},
\label{eq:clusterposterior}
\end{equation}
where $\tau$ is temperature parameter that controls the variance of the distribution. As we decrease the temperature, the distribution becomes more peaky, and becomes flatter as we increase it.
The importance of $\tau$ can be understood by studying the loss function $L_{\phi}$ in line 15 of Algo.~\ref{algo:episodic_training}. During training, if clusters are well-separated, $q(z|k,x)$ will be peaky so that each example effectively contributes to the update of a single cluster in a class, whereas if clusters overlap, $q(z|k,x)$ will be diffuse and the corresponding example will contribute to multiple prototypes.  Therefore, during training, we typically set $\tau$ to favor peaky distributions so that learned clusters focus on different regions of the input space. 





\noindent\textbf{Class-specific cluster prototypes:} 
In episodic training, an epoch corresponds to a fixed number of episodes and within each episode, classes are sampled uniformly.
In our setting with huge class imbalances, this translates to  examples from tail classes being oversampled, while examples from the head may be undersampled within an epoch. This can adversely affect model training.
To mitigate this, at the start of an epoch, we initialize cluster prototypes for each class using k-means on the learned embedding representation of examples from the entire training set of that class. We rerun this clustering step at the start of each epoch to prevent collapse to using only a single cluster per class.

Subsequently, in each episode, we use an \textit{online} update scheme that balances between the local estimate of the prototype computed from embeddings of the current support set (to account for the evolving embedding space), and the prototypes learned so far:
\begin{equation}
 \mu_{z,k}^{new}  \leftarrow \alpha \mu_{z,k}^{old} + (1-\alpha) \frac {\sum_{(x,y)\in S_k} q(z | k,x) f_\phi(x)} {\sum_{(x,y)\in S_k} q(z | k,x)}
 \label{eq:prototype_update},
\end{equation} 
where $\alpha$ trades off memory from previous episodes and its current estimate. 

\vspace{-5pt}
\subsection{Understanding the role of multiple clusters}
\label{sec:linear}

We can derive insights about the role of multiple clusters by interpreting PCN as a non-linear generalization of PN (\cite{snell2017prototypical}). Using squared Euclidean distance in Eqn.~\ref{eq:pyGx}, we expand the term in the exponent so that:
\begin{IEEEeqnarray}{rCl}
\IEEEeqnarraymulticol{3}{l}{
-\sum_{z} q(z | k, x) ||f_\phi(x) - \mu_{z,k}||^2 
}\nonumber\\ \hspace{-1in}
& = & -\sum_{z} q(z | k, x) f_\phi(x)^{T}f_\phi(x) -\sum_{z} q(z | k, x) \mu_{z,c}^{T}\mu_{z,k}
\nonumber \\ 
&& \negmedspace {} + 2 \sum_{z} q(z | k, x)f_\phi(x)^{T}\mu_{z,k}\\
& = & \textrm{constant for k} +  w_{k,x}^{T} f_\phi(x) - b_k 
\label{eqn:expansion}
\end{IEEEeqnarray}
where
\vspace{-20pt}
\begin{eqnarray}
w_{k,x} &=& 2  \sum_{z} q(z | k, x) \mu_{z,k} \\
b_{k,x} &=& \sum_{z} q(z | k, x) \mu_{z,k}^{T}\mu_{z,k} 
\end{eqnarray}
The last two terms in 
Eqn.~\ref{eqn:expansion} are non-linear functions of the data, where the non-linearity is captured through both the embedding and the mixing variables. The functional forms of the factors, namely $w_{k,x}$ and $b_{k,x}$, also sheds light on the advantage of using multiple clusters per class.  In particular, unlike in prototypical networks, $w_{k,x}$ is an \emph{example-specific} ``prototypical'' representation for class $k$, obtained by using a convex combination of prototypes for the class, weighted by posterior probability over within-class cluster assignments. When $q(z | k, x)$ is confident with a peaky posterior, the model behaves like a regular prototypical network. In contrast, when the posterior has uncertainty, PCN interpolates between the prototypes by modulating $q(z | k, x)$.  

%% file: source/experiments.tex
\section{Results}

\subsection{Experimental setup}

\noindent\textbf{Dataset:}  
We construct our dataset from the Dermnet Skin Disease Atlas\footnote{http://www.dermnet.com/}, one of the largest public photo dermatology sources containing over 23,000 images of dermatological conditions. Images are annotated at a two level hierarchy -- a coarse top-level containing parent 23 categories, and a fine-grained bottom-level containing more than 600 skin conditions. 
We focus on the more challenging bottom-level hierarchy for our experiments. 
First, we remove duplicates from the dataset based on name, and also based on collisions found using perceptual image hashing~\cite{zauner2010implementation}.

Figure~\ref{fig:teaser} presents a histogram of the resulting class distribution, filtered to the top-200 classes. We can see that the dataset has a long tail with only the 100 largest classes having more than 50 images; beyond 200 classes, the number of images per class reduces to double digits, and with 300 classes to single digits. Unless otherwise stated,  for experimental comparisons, we focus on the top-200 classes so that $K_{base+novel} = 200$, which contains $15507$ images. Similar to \cite{hariharan2017low}, we treat the largest 150 classes as base classes ($K_{base}=150$) and the remaining 50 classes as novel ($K_{novel}=50$). This helps in ensuring  reasonably sized splits for training, validation and  evaluation.  In particular, we sample $max(5, 20\%)$ without replacement for each base class to get validation and test splits ($3163$ images each). The remaining is used for training ($9181$ images). For the low-shot learning phase, following the procedure used in \cite{hariharan2017low}, we sample 5 examples each for training and testing, respectively. We report mean and standard deviation of metrics over 10 cross validation runs.


\noindent\textbf{Metrics:} We report mean of per-class accuracy ($\textrm{mca}$), treating each class as equally important. For a dataset consisting of $C$ classes, with  $T_c$ examples in each class,  mean accuracy is the average of per-class accuracies:
\begin{equation}
\textrm{mca} = \frac{1}{C}\sum_{c}\frac{\sum_{t=1}^{T_c} I[\hat{y}^{(t)}[0] = y^{(t)}]}{T_k},
\end{equation}
where, for $t^{th}$ example, $\hat{y}^{(t)}[j]$ is the $j^{th}$ top class predicted from a model and $ y^{(t)}$ is its corresponding ground truth label, where $I$ denotes the indicator function.

We use $\textrm{mca}_{\textrm{base+novel}}$ to report combined mca performance of examples from all classes.  $\textrm{mca}_{\textrm{base}}$ corresponds to evaluation of classifier on $K_{base+novel}$ classes but restricted to only test examples from base classes. Similarly, $\textrm{mca}_{\textrm{novel}}$ corresponds to evaluation of test examples in novel c{lasses while performing $K_{base+novel}$ way classification. 

We also report recall@k (k $\in \{5,10\}$). 
This metric (also called {\emph{sensitivity} in the medical literature) is valuable in deployment contexts that involve aiding doctors in diagnosis. This metric is not as strict as $\textrm{mca}$ but it ensures that the relevant disease condition is considered within a small range of false positives. 

\begin{table*}
\centering
\begin{tabular}{c|ccc|ccc} 
\toprule
\multicolumn{1}{c|}{} & \multicolumn{3}{c|}{$n = 5$}                                                                                                                                                         & \multicolumn{3}{c}{$n = 10$}                                                                                                                                                  \\
Approach                  & \begin{tabular}[c]{@{}c@{}}$\textrm{mca}_{\textrm{base+novel}}$\\ \end{tabular} & \begin{tabular}[c]{@{}c@{}}$\textrm{mca}_{\textrm{base}}$\\ \end{tabular} & \begin{tabular}[c]{@{}c@{}}$\textrm{mca}_{\textrm{novel}}$ \end{tabular} & \begin{tabular}[c]{@{}c@{}}$\textrm{mca}_{\textrm{base+novel}}$\end{tabular} & \begin{tabular}[c]{@{}c@{}}$\textrm{mca}_{\textrm{base}}$\end{tabular} & \begin{tabular}[c]{@{}c@{}}$\textrm{mca}_{\textrm{novel}}$\end{tabular}  \\ 
\cmidrule{1-7}
$FT_{150}$-1NN                    & 46.18 +/- 0.81                                                & 55.32 +/- 0.30                                        & 18.76 +/- 3.30                                        & 49.51 +/- 0.34                                                       & 54.86 +/- 0.50                                                 & 33.44 +/- 1.35                                                   \\
$FT_{150}$-3NN                    & 44.28 +/- 0.32                                                & 54.77 +/- 0.47                                        & 12.80 +/- 1.50                                        & 47.01 +/- 0.56                                                       & 54.13 +/- 0.43                                                 & 25.64 +/- 1.51                                                   \\ 
$FT_{200}$-1NN                    & 46.52 +/- 0.39                                                & 54.17 +/- 0.30                                        & 22.50 +/- 0.75                                        & 49.92 +/- 0.47                                                       & 53.80 +/- 0.35                                                 & 38.27 +/- 1.32                                                   \\ 
$FT_{200}$-3NN                    & 44.69 +/- 0.39                                                & 52.61 +/- 0.21                                        & 20.93 +/- 2.00                                        & 47.96 +/- 0.11                                                       & 52.53 +/- 0.14                                                 & 34.27 +/- 0.19                                                   \\ 
$FT_{200}$-CE                     & \textbf{47.82 +/- 0.46}                                        & \textbf{55.75 +/- 0.71}                               & 24.00 +/- 3.22                                            & \textbf{51.51 +/- 0.41}                                              & \textbf{55.21 +/- 0.26}                                       & 40.40 +/- 2.36                                          \\
\cmidrule{1-7}
PN                        & 43.92 +/- 0.40                                          & 48.71 +/- 0.37                                    & 29.56 +/- 2.35                                     & 44.93 +/- 0.79                                            & 47.55 +/- 0.37                                      & 37.08 +/- 3.39                 \\
PCN (ours)                       & \textbf{47.79 +/- 0.71}                                       & 53.70 +/- 0.18                                        & \textbf{30.04 +/- 2.77}                                & \textbf{50.92 +/- 0.63}                                            & 51.38 +/- 0.34                                      & \textbf{49.56 +/- 2.76}                                        \\ 
\bottomrule
\end{tabular}\caption{Mean per-class accuracy (MCA) on top 200 classes. We focus on the low-shot setting, using all training data for the base classes (the largest 150) and  $n=5$ or $10$ examples for the remaining 50 classes (denoted as ``novel"). Note that $FT_{200}$-CE and $FT_{200}$-*NN use training data for all 200 classes, whereas the other models use only the base classes for representation learning, using support sets from the remaining 50 classes after training to derive prototypes. We report recall metrics in the appendix.}
\label{tab:main}
\end{table*}

\noindent\textbf{Model:} We initialize a 50-layer ResNet-v2 convolutional neural network~\cite{he2016deep} with ImageNet pretraining, and train a Prototypical Clustering Network as described in Sec.~\ref{sec:approach} on $K_{base}$ classes. We use 10 and 4 clusters per class for base and novel classes respectively (picked via grid search). 

\noindent\textbf{Baselines}
\begin{itemize}
\item  Prototypical Network (PN): We train an ImageNet Pretrained ResNet-V2 CNN as a Prototypical Network \cite{snell2017prototypical} on $K_{base}$ classes.

\item Finetuned Resnet with nearest neighbor ($FT_{K}$-*NN):  Here, we finetune an ImageNet-pretrained ResNet-v2 convolutional neural network with 50 layers \cite{he2016deep} on training data from $K$ classes. We report numbers for K $\in \{K_{base}, K_{base+novel}\}$. The model is trained as a softmax classifier with a standard cross entropy objective.   Then, we obtain embeddings for the entire training set consisting of $K_{base+novel}$ classes. This is used to perform $*$-nearest neighbor classification on the test set from all of  $K_{base+novel}$ classes.

\item Finetuned ResNet ($FT_{K}$-CE):  We use the same ResNet model as the above, with $K = K_{base+novel}$, i.e. trained for $K_{base+novel}$ way classification using training data from both base and novel classes, and validated using the corresponding validation set on the base classes (due to lack of data in novel classes).  We train the model with class balancing.  This is a strong baseline as we use all $K_{base+novel}$ during training, and also due to class balancing, which has shown to improve generalization \cite{buda2018systematic}.
\end{itemize}

\noindent\textbf{Hyperparameters:} For PN and PCN,  we use episodic batching with 10-way 10-shot classification (at train), and 200 episodes per epoch. At test, we compute per-class prototypes using the training set for all $K_{base+novel}$ classes, and perform $K_{base+novel}$ way classification. The embedding function for PCN and PN produces 256-dimensional embeddings, and uses the same architecture as in $FT_{K}$-CE (with one less fully connected layer). Models are trained with early stopping using Adam~\cite{kingma2014adam}, a learning rate of $10^{-4}$, and L2 weight decay of $10^{-5}$.

\subsection{Main results}
Table~\ref{tab:main} highlights our main results. The table shows test set MCA over the 200 classes available during test time for two different low shot settings: train shots of 5 and 10 with test set of 5. In both low shot settings, we observe the following trends:
\begin{itemize}
\item  $FT_{K}$-CE and PCN shares similar performance on combined MCA. However, their performance on the base and novel classes are quite distinct. Much of the performance gains for $FT_{K}$-CE come from the base classes that have a lot more training examples than novel classes. In contrast, PCN, through episodic training aims at learning discriminative feature representations that are generalizable to novel classes with highly constrained number of examples; this is evident by its significantly better performance (9\% absolute gains) in generalizing to novel classes. At the same time, PCN ensures that performance on novel classes doesn't come at the cost of lower accuracy on the base classes. Also note that the FT-CE model requires re-training for adding novel classes while PCN only requires a single forward pass to learn prototypes for novel classes. 

\item $FT_{K}$-*NN models learn robust representations for base classes, but are unable to generalize to novel classes, outperforming a regular PN model on top-200 MCA but underperforming against PCN. Interestingly, we find that increasing the number of nearest neighbors leads to poor performance, especially on novel classes. This could be due to sparsity of training data. 

\item PCN outperforms PN on combined base and novel classes by a large margin. This demonstrates that representing classes with multiple prototypes leads to better generalization on both base and novel classes.  In Figure~\ref{fig:prototypeNN}, we show the nearest neighbor to class prototype for PN and to four of the PCN prototypes, for select classes. We can see that PCN has learned to model intra-class variability much more effectively. As an example, for eczema  and acne classes we can see that PCN learns clusters corresponding to these skin conditions in different anatomical regions. We  provide a more in-depth comparison in the next section. 
\end{itemize}
\begin{figure}[t]
 \centering 
 \includegraphics[width=\linewidth]{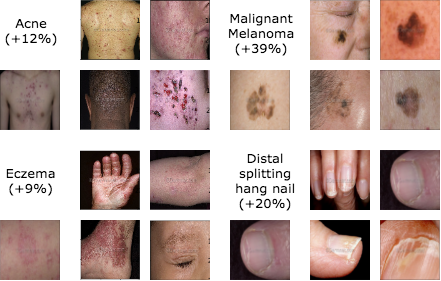}
 \vspace{-10pt}
 \caption{Learned prototypes are shown using their nearest neighbors in the training set. Each skin condition is in a $2 \times 3$ grid; The image below the name of the skin condition corresponds to PN while the $2\times 2$ grid corresponds to nearest neighbors of four cluster prototypes. +X\% below the name denotes improvement of PNC over PN for that class for $\textrm{mca}_{base+novel}$,. Note that novel classes such as `Distal splitting hang nail' can also be diverse, as shown by clusters identified with PCN. }
\vspace{-13pt}
\label{fig:prototypeNN}
\end{figure}

\begin{table*}
\centering
\resizebox{0.9\textwidth}{!}{%
\begin{tabular}{ccccccc} 
\toprule
Model & \begin{tabular}[c]{@{}c@{}}Eval CPC\\ (base / novel) \end{tabular} & \begin{tabular}[c]{@{}c@{}}$\textrm{mca}_{\textrm{base+novel}}$\end{tabular} & \begin{tabular}[c]{@{}l@{}}$\textrm{mca}_{\textrm{base}}$\end{tabular} & \begin{tabular}[c]{@{}l@{}}$\textrm{mca}_{\textrm{novel}}$\end{tabular} & recall@5              & recall@10              \\ 
\cmidrule{1-7}
PN    & 1 / 1                                                              & 43.92 +/- 0.40                                          & 48.71 +/- 0.37                                    & 29.56 +/- 2.35                                      & 70.88 +/- 0.36           & 80.19 +/- 0.26            \\
PN    & 1 / 4                                                              & 44.35 +/- 0.53                                          & 50.35 +/- 0.42                                    & 26.36 +/- 2.34                                      & 74.16 +/- 0.21           & 83.45 +/- 0.25            \\
PN    & 10 / 4                                                             & 43.78 +/- 0.78                                          & 50.30 +/- 0.21                                    & 24.20 +/- 3.02                                      & 75.58 +/- 0.23           & 84.03 +/- 0.19            \\ 
\hline
PCN (ours)  & 10 / 4                                                             & \textbf{47.79 +/- 0.71}                                 & \textbf{53.70 +/- 0.18}                                    & \textbf{30.04 +/- 2.77}                                     & \textbf{77.76 +/- 0.19}  & \textbf{85.96 +/- 0.38}   \\
\bottomrule
\end{tabular}
}
\caption{Does post-hoc clustering on PN help?}\label{tab:eval_ablations}
\end{table*}

\subsection{Comparison between PCN and PN}
\label{sec:pcnVSpn}
\noindent\textbf{ PCN or PN with post-hoc clustering?}
To understand the effectiveness of PCN, we compare it to a PN model in which we perform  ``post-hoc'' clustering: (a) cluster novel class representations using the PN model's learned embeddings (with cluster size of 4) (b) cluster both base and novel class representations (with cluster size of 10 and 4, as in PCN). Rows 1-3 in Table~\ref{tab:eval_ablations} compare the performance between different post-hoc clustering variants of PN against PCN. We see that PCN leads in all metrics across the board; thus such post-hoc clustering does not lead to improved performance. A reason for this is that the PN model is optimized to learn representations assuming a projection to a single cluster for each class, and hence clustering on such learned representation does not improve performance. This further validates the importance of training with multiple clusters.


\begin{figure}[t]
 \centering 
 \includegraphics[width=0.7\linewidth]{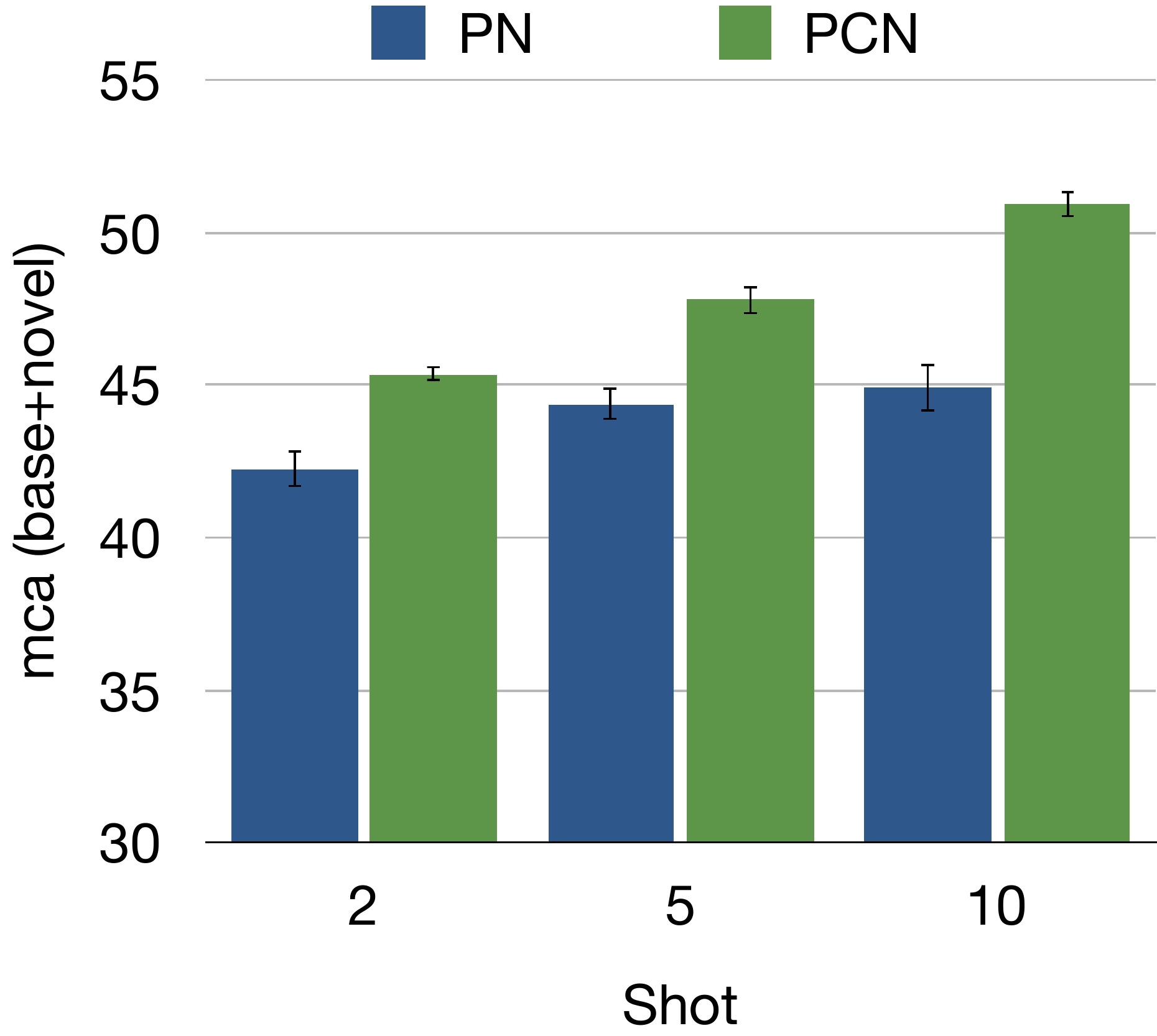}
 \vspace{-10pt}
 \caption{
 Comparison between PN and PCN as a function of training shot size for novel classes.}
\vspace{-13pt}
\label{fig:progressive}
\end{figure}

\noindent\textbf{Role of shot in novel classes:} Figure~\ref{fig:progressive} highlights the effect of number of support examples (shot). As we increase the shot, the performance improves on both methods, but that improvement is larger for PCN than for PN. Because of this, the performance gap between the two methods drastically increases. PCN is better at utilizing the availability of more data by partitioning the space with clusters.

\noindent\textbf{Effect of increasing the novel classes:}
In this experiment, we study the performance as we vary the number of novel classes at test time from 50 to 150, bringing the total number of classes up from 200 to 300.  Fig~\ref{fig:extension} provides the comparison. We used a train and test shot of 2 and 5, respectively since most classes in these additional 100 novel classes in the long tail have less than 10 examples. Results are reported with 10-fold cross validation. While there is a drop in performance for both models due to very small shot sizes, we can see that the performance gap between PCN and PN continues to hold.

\begin{figure}
 \centering 
 \includegraphics[width=0.8\linewidth]{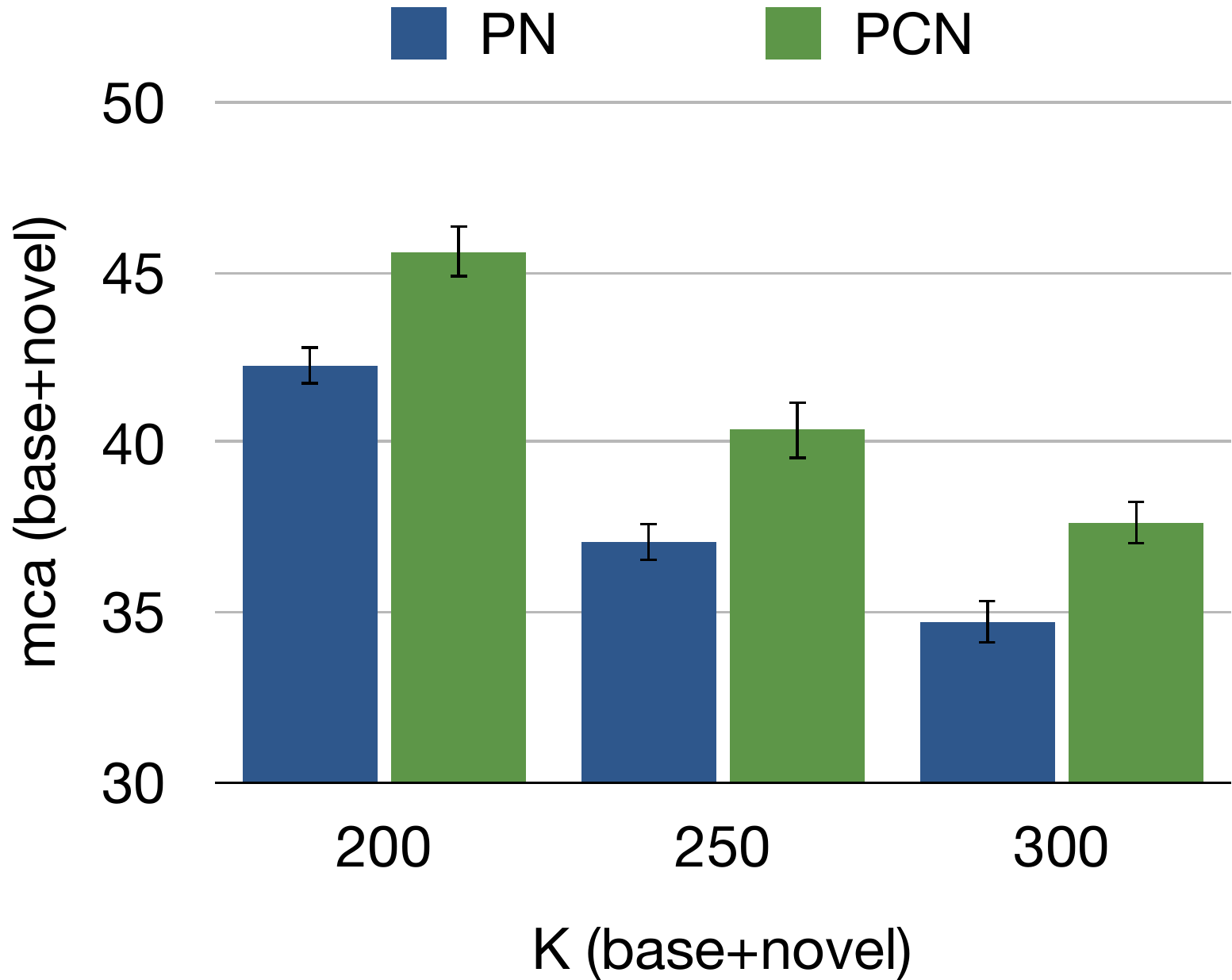}
 \vspace{-10pt}
 \caption{Comparison between PN and PCN as a function of number of novel classes. Due to lack of sufficient data, we compare using a train shot of 2 and test shot of 5.}
\vspace{-13pt}
\label{fig:extension}
\end{figure}

\begin{figure}
 \centering 
 \includegraphics[width=0.7\linewidth]{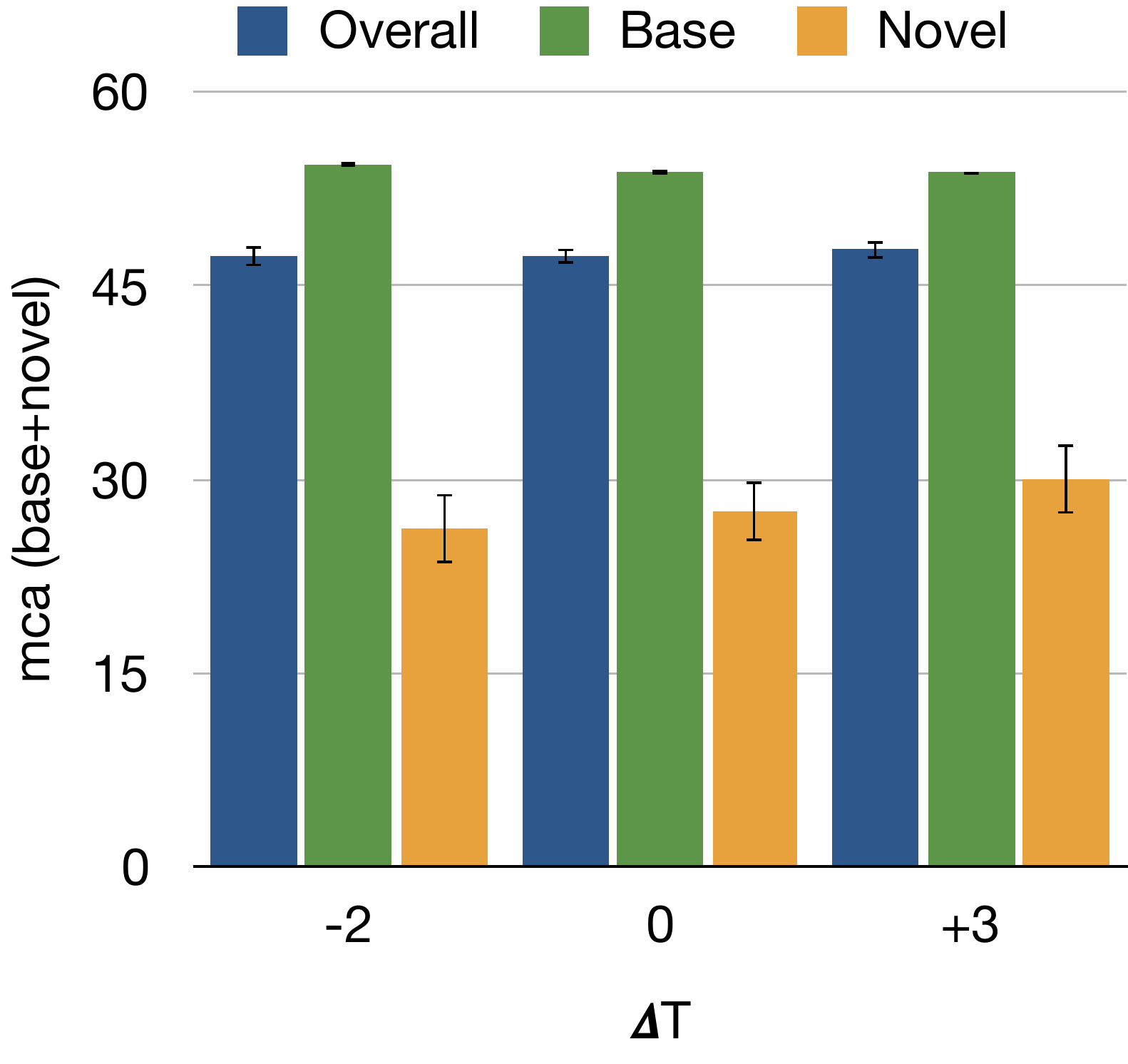}
 \vspace{-10pt}
 \caption{
Effect of temperature on PCN}
\vspace{-13pt}
\label{fig:temperature}
\end{figure}
\vspace{5pt}
\begin{table}
\centering
\caption{Importance of episodic memory}
\begin{tabular}{ccc} 
\toprule
Approach & $\alpha$  & \begin{tabular}[c]{@{}c@{}}$\textrm{mca}_{\textrm{base+novel}}$ \end{tabular}  \\ 
\midrule
PCN      & 0         & 45.62 +/- 0.89                                                         \\
PCN      & 0.5       & 47.49 +/- 0.71                                              \\
\cmidrule{1-3}
PN       & 0         & 44.35 +/- 0.53                                              \\
PN*      & 0.5       & 45.84 +/- 0.46                                                         \\
\bottomrule
\end{tabular}
\vspace{-10pt}
\label{tbl:alpha}
\end{table}

\begin{figure*}[t]
 \centering 
 \includegraphics[width=0.7\textwidth]{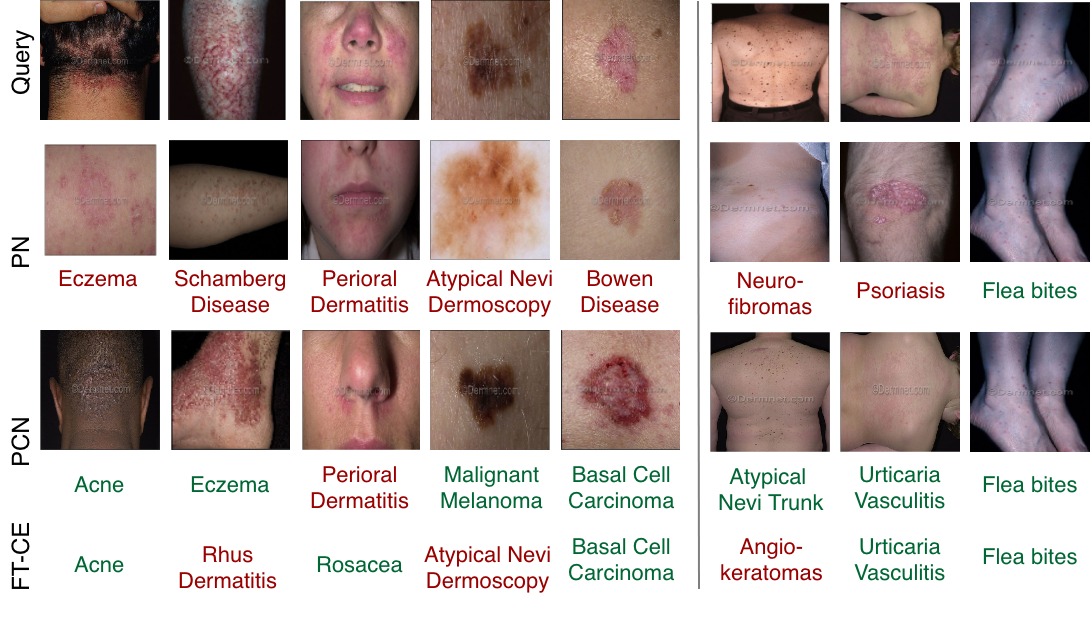}
 \vspace{-10pt}
 \caption{For each query image in test set, we compare PCN with PN and $FT_{200}$-CE.  For each image, we color code correct label with green and incorrect with red. For PN, we show the nearest neighbor to the prototype of the \emph{predicted} class. For PCN, we show the nearest neighbor of the top cluster according to $q(z|c,x)$  of the predicted class. The last three columns correspond to examples from novel classes.}
\vspace{-13pt}
\label{fig:qualitative_v0}
\end{figure*}

\subsection{Role of Hyperparameters}
\label{sec:hyperparameters}
\noindent\textbf{Importance of Temperature:} 
Fig.~\ref{fig:temperature} presents the performance by varying  $\Delta_\tau = \tau_{test}-\tau_{train}$, the difference in temperature used in test versus train time. We can see that the performance is agnostic for the base classes, as these classes have been used in the training phases. However, for the novel classes, higher interpolation through an increased temperature leading to  $\Delta_\tau > 0$ leads to improved performance. Conversely, when $\Delta_\tau < 0$, performance drops. This means that at test time, the model requires interpolating between the cluster prototypes to effectively predict a class label, as described in Sec~\ref{sec:linear}.


\noindent\textbf{Does episodic memory help?}  Table~\ref{tbl:alpha} shows that we can get improvements even with a simple online update rule that blends prototypes computed using the support set in the current episode with the past, using $\alpha=0.5$. This trend is also seen for prototypical networks (denoted by PN*). We leave as future work the task of modeling adaptive $\alpha$. 

\begin{figure*}
 \centering 
 \includegraphics[width=0.8\linewidth]{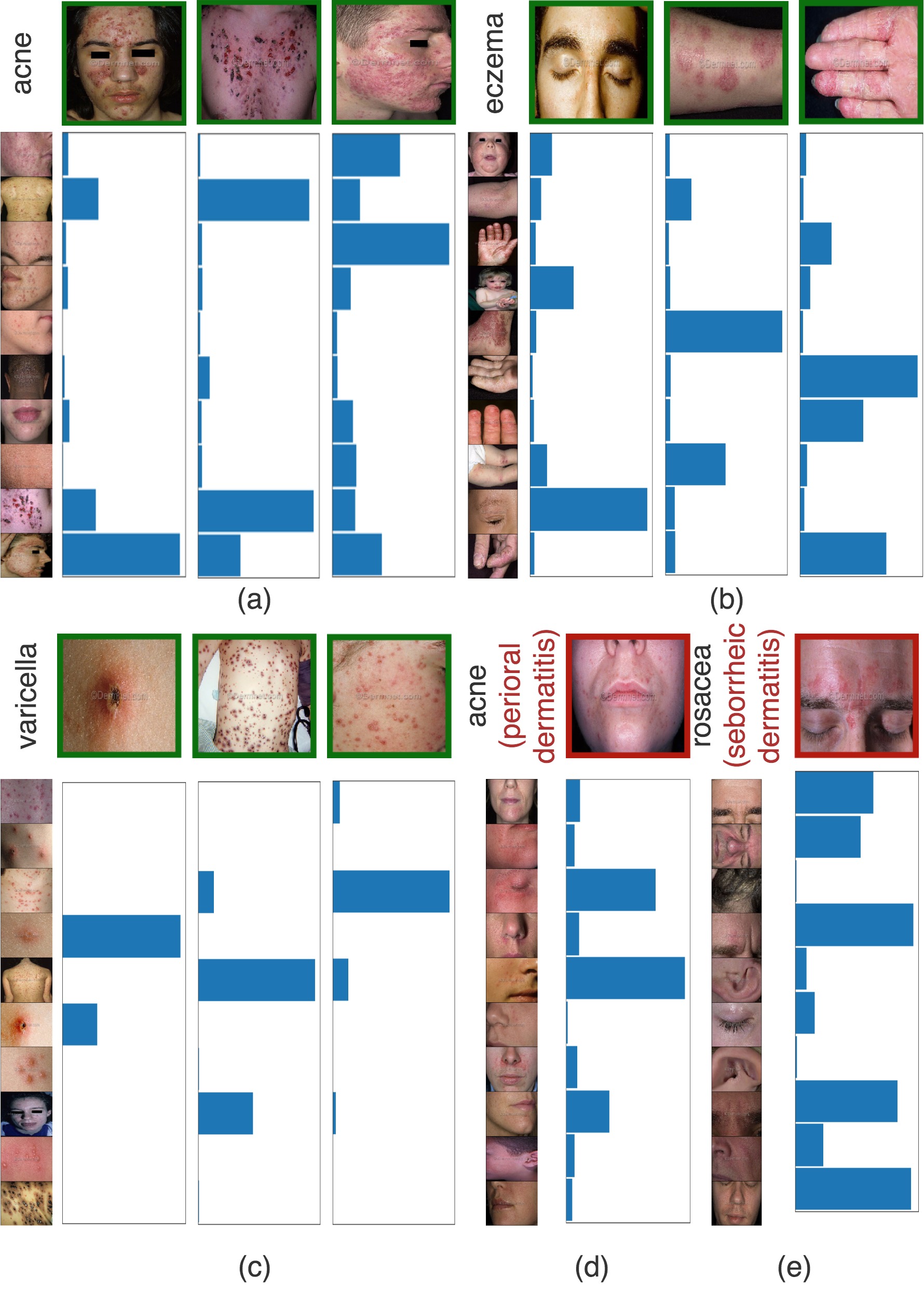}
 \caption{Effectiveness of using multiple clusters.  Shown for base classes. (a)-(c): Examples from test set that are correctly classified by PCN. For each class, we show the nearest neighbor to the learned prototypes. We also present three examples (columns) whose labels are correctly predicted and the inferred cluster responsibilities $q(z|c,x)$ conditioned on the correct class. (d)-(e): Examples from the test set that are incorrectly classified by PCN. Correct label is shown in black, while the incorrect prediction is shown in red. We show the nearest neighbors to the learned cluster prototypes of the \textit{predicted} (incorrect) class, and the corresponding cluster responsibilities. Note that green outlines around query images denote correct classification while red denotes incorrect classification.}
\vspace{-13pt}
\label{fig:interpolate}
\end{figure*}

\subsection{Qualitative Results}
Figure~\ref{fig:qualitative_v0} provides qualitative examples comparing the three methods. 
Acne is one of the largest classes in the base classes with large intra-class variability. Both FT-CE and PCN can diagnose this example correctly. However, PN due to its limited capacity to represent the huge variability in the class is confused with another large class, namely, eczema. PCN, due to having access to multiple clusters can learn a better representation and correctly diagnose acne.

In column 4, we present a case in which both PN and FT-CE identified the query image as atypical nevi dermoscopy, while PCN correctly classified it as malignant melanoma. Atypical nevi are `funny-looking' moles that are precursors to melanoma. It has been recently studied that dermoscopic features discriminating between atypical naevi and melanoma require expert interpretation through longitudinal monitoring, but are often ignored as simple moles \cite{Stanley07}. In contrast, consider Column 7:  FT-CE misdiagnose atypical nevi as Angiokeratoma, a benign skin lesion of capillaries, resulting in small marks of red to blue color. In the data-starved setting, FT-CE and PN are unable to differentiate the two skin conditions while PCN can better match up to the support set. 

\begin{figure*}
 \centering 
 \includegraphics[width=0.85\linewidth]{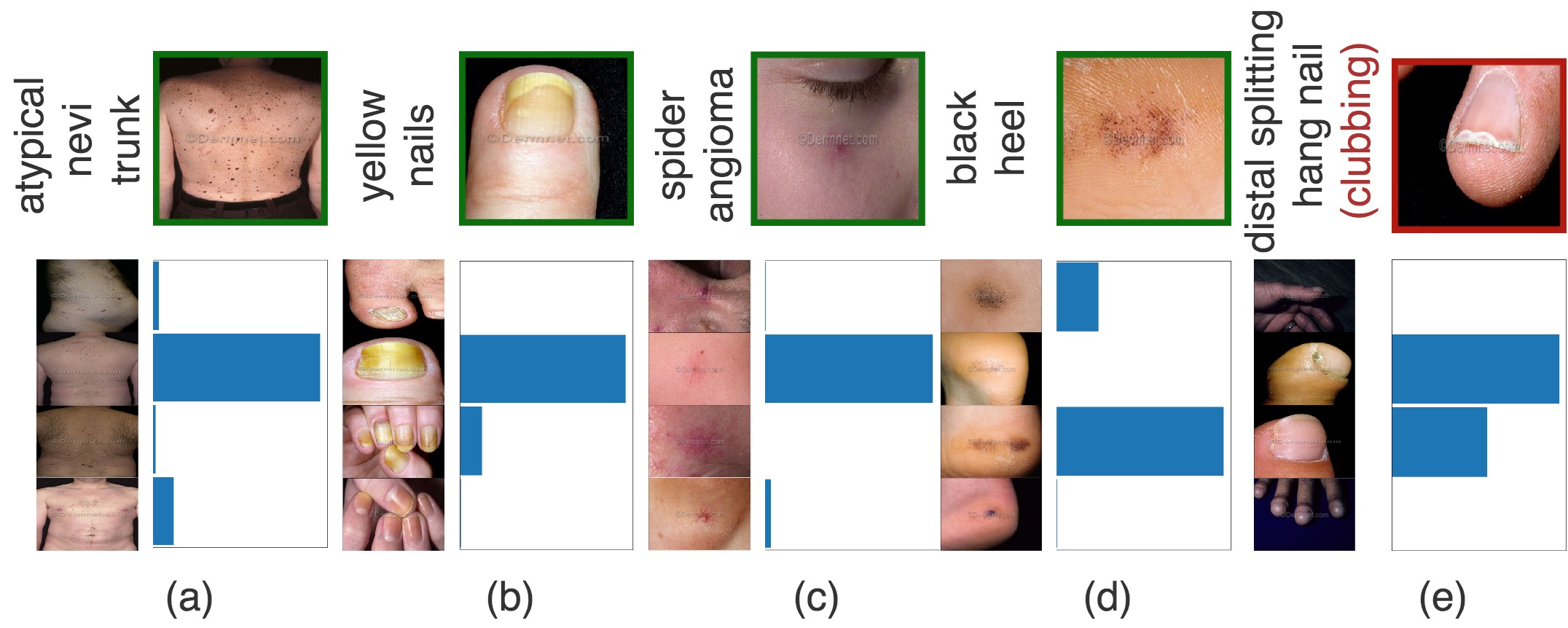}
 \caption{Effectiveness of using multiple clusters for novel classes. (a)-(d) examples from test set that are correctly classified by PCN. For each class, we show the nearest neighbor to the prototypes of the four clusters that are learned for each novel class. For each example, the distribution of inferred cluster responsibilities $q(z|c,x)$ conditioned on the correct (predicted) class is shown. (e): Query example that is incorrectly classified by PCN. Correct label is shown in black, while incorrect prediction is in red. Shown alongside are the nearest neighbors to the cluster prototypes of the \textit{predicted} (incorrect) label, and the corresponding cluster responsibilities.}
\vspace{-13pt}
\label{fig:interpolate_novel}
\end{figure*}

\par \noindent
\textbf{Effectiveness of multiple clusters}. In Sec~\ref{sec:approach}, we show how PCN can interpolate between the learned prototypes by modulating $q(z|c,x)$. In Figure~\ref{fig:interpolate} we show some qualitative examples to illustrate this. We show query images from the test set for various classes, with a mix of correct and incorrectly classified examples. Below the class label, we show the nearest neighbor image from the training set to each of the learned prototypes for the class \textit{predicted} by PCN, and below each query image, cluster responsibilities placed by the model on each of these prototypes.
As an example, consider examples corresponding to acne in 
Figure~\ref{fig:interpolate}(a). For this class, we show three examples, all of which are correctly classified by PCN.  We can see that the $q(z|c,x)$ distribution varies quite a bit across examples, being a lot more diffuse in some cases than others. It can also be seen that the model learns to accurate place probability mass on similar prototypes. For instance, in column 2 of ~\ref{fig:interpolate}(a), the model appears to interpolate between two prototypes that are similar to the query image in pose and skin texture respectively, to make a correct prediction. Similarly for ~\ref{fig:interpolate}(b) (eczema), the model is accurately able to identify eczemas on the face, arm, and hand, by combining the most relevant prototypes. While these classes see relatively diffuse responsibility distributions, the distribution is far more peaked for the varicella class in (c). Figure~\ref{fig:interpolate}(d)-(e) shows incorrectly predicted examples; Even for these examples, the model seems to interpolate, albeit incorrectly, to make predictions.
In Figure~\ref{fig:interpolate_novel}, we show similar examples for novel classes. 

%% file: source/conclusion.tex

\section{Discussion}
\par
We now discuss some of the salient features of our approach that make it well suited to deployment on healthcare platforms. First, our approach is \textit{privacy preserving}, i.e. it is trivial to extend a base classifier (possibly trained on large hardware infrastructures and deployed as a service) to learn to recognize novel conditions, without requiring access to either the original (potentially proprietary) training data, or training infrastructure (a single forward pass is required). 
\par
Additionally, our approach makes it possible to very quickly generate learning curves for novel classes, which can be used to guide data acquisition for classes which are determined most likely to benefit from additional data (active few-shot learning). On the other hand, re-training FT-CE to create such learning curves would be extremely time and computationally intensive.
\par
While in this work we focus the applicability of our approach in few-shot diagnosis, there exist a number of future directions worth pursuing. The true effectiveness and utility of our system is in aiding the physician, and this requires studies that include such a deployment. Immediate extensions include determining the absence of any condition (adding i.e. a `normal' class), and controlling for demographic variables when appropriate. Another interesting direction would be to incorporate additional modalities of data for more robust prediction. Dermatologists use symptoms that patients experience, such as itchiness of the skin, in disambiguating skin conditions \cite{Resneck16}. Incorporating these medical symptoms as part of the classification task will be an interesting direction to pursue. Finally, while this approach has been developed in the context of the very specific needs of dermatological diagnosis, we believe that similar requirements exist in other domains. As follow up work, we will study the generalization of our approach across domains and datasets.

\vspace{-5pt}
\section{Conclusion and Future Work}
\vspace{-5pt}
We propose Prototypical Clustering Networks: a few shot learning approach to dermatological image classification. This method is scalable to novel classes, and can effectively capture intra-class variability. We observe that our approach outperforms strong baselines on this task, especially on the long tail of the data distribution. Such a machine learning system can be a valuable aid to telemedicine services, and improve access, equity, quality, and cost-effectiveness of healthcare.



%% file: source/appendix.tex
\appendix

\section{Appendix}

\par \noindent
In this appendix we first provide a per-class performance comparison of our proposed approach against the FT-CE baseline. Next, we report additional recall@k metrics for all studied approaches.

\begin{figure*}
 \centering 
 \includegraphics[width=0.5\linewidth]{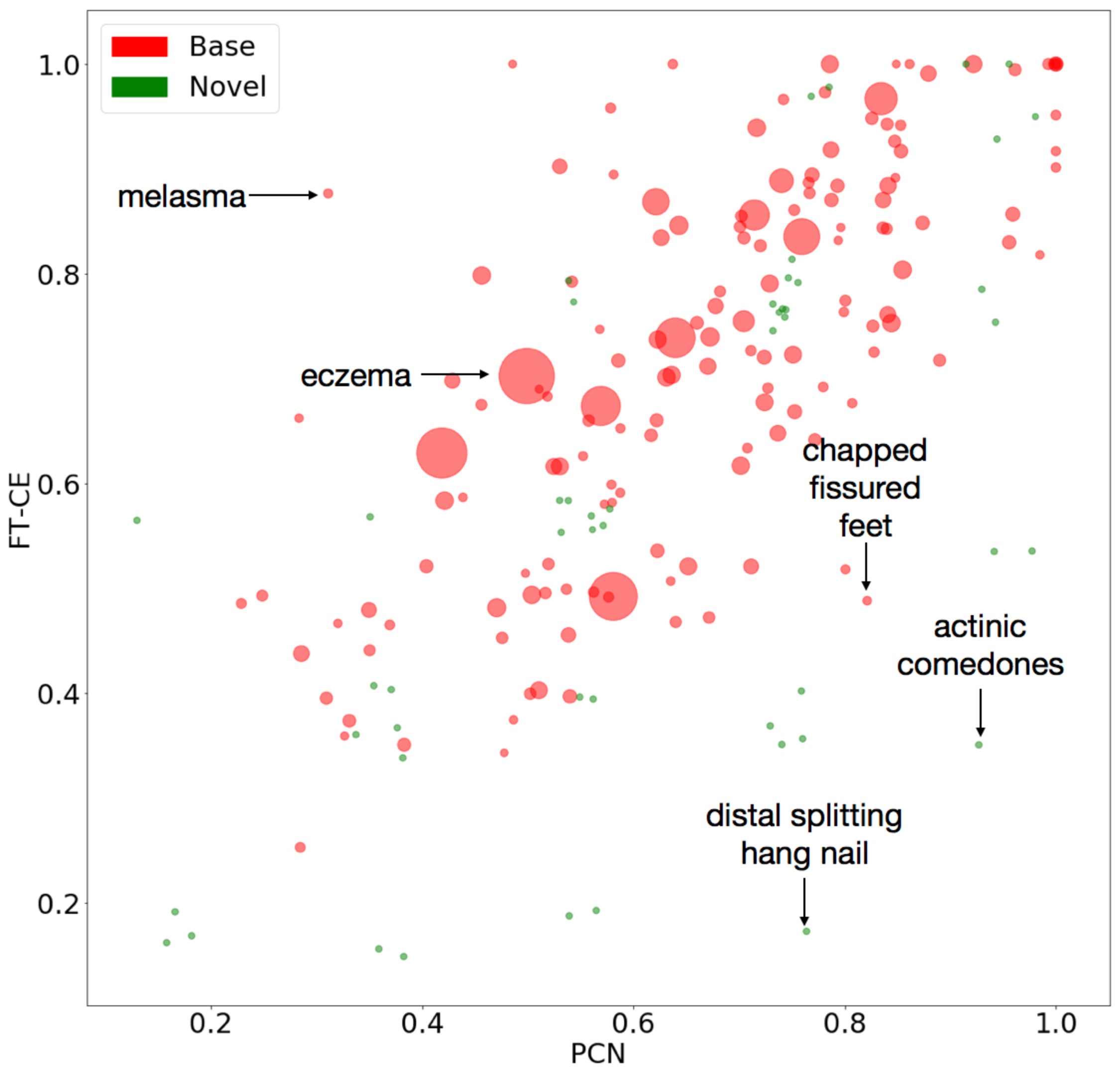}
 \vspace{10pt}
 \caption{
Comparison between $FT_{200}$-CE and PCN: Per-class accuracy. Each class is denoted by a dot and the area of dot is proportional to the number of training examples for the class.}
\label{fig:scatter}
\end{figure*}

\begin{table*}
\singlespace
\centering
\tabcolsep=0.11cm

\caption{Recall@k on top 200 classes. 
}
\begin{tabular}{cccccccc} 
\toprule
                      & Approach                                              & $\textrm{r@5}_{\textrm{base+novel}}$ & $\textrm{r@5}_{\textrm{base}}$ & $\textrm{r@5}_{\textrm{novel}}$ & $\textrm{r@10}_{\textrm{base+novel}}$ & $\textrm{r@10}_{\textrm{base}}$ & $\textrm{r@10}_{\textrm{novel}}$  \\ 
\hline
\multirow{3}{*}{n=5}  & FT-CE                                                 & 77.7 +/- 0.79                                                           & 80.84 +/- 0.83~                                                       & 38.0 +/- 2.14~                                                         & 84.92 +/- 0.47                                                               & 88.06 +/- 0.44                                                         & 45.33 +/- 1.47                                                           \\
                      & PN                                                    & 70.88 +/- 0.36                                                          & 71.48 +/- 0.33~                                                       & 63.24 +/- 2.34~                                                        & 80.19 +/- 0.26                                                               & 80.50 +/- 0.18                                                         & 76.28 +/- 2.13                                                           \\
                      & PCN (ours)                                            & 77.76 +/- 0.19                                                          & 79.23 +/- 0.22                                                        & 59.24 +/- 2.57                                                         & 85.96 +/- 0.38                                                               & 87.05 +/- 0.20                                                         & 72.16 +/- 3.59                                                           \\ 
\hline
\multirow{3}{*}{n=10} & \multicolumn{1}{c}{FT-CE}                             & 78.59 +/- 0.13                                                          & 80.22 +/- 0.33                                                        & 58.0 +/- 3.22~                                                         & 86.42 +/- 0.25                                                               & 88.05 +/- 0.12                                                         & 65.87 +/- 2.62                                                           \\
                      & \multicolumn{1}{c}{PN}                                & 69.36 +/- 0.29                                                          & 69.14 +/- 0.31                                                        & 72.16 +/- 1.66                                                         & 78.59 +/- 0.25                                     & 78.11 +/- 0.26                                                         & 84.68 +/- 2.22                                                           \\
                      & \multicolumn{1}{c}{PCN (ours)} & 76.29 +/- 0.22                                                          & 76.43 +/- 0.23                                                        & 74.52 +/- 2.62                                                         & 85.03 +/- 0.23                                                               & 85.04 +/- 0.25                                                         & 85.04 +/- 2.00                                                           \\
\bottomrule
\end{tabular}
\label{tab:main}
\end{table*}
\begin{table*}
\singlespace
\centering
\tabcolsep=0.1cm
\caption{Balanced Recall@k on top 200 classes. 
}
\begin{tabular}{cccccccc} 
\toprule
                      & Approach                                              & $\textrm{br@5}_{\textrm{base+novel}}$ & $\textrm{br@5}_{\textrm{base}}$ & $\textrm{br@5}_{\textrm{novel}}$ & $\textrm{br@10}_{\textrm{base+novel}}$ & $\textrm{br@10}_{\textrm{base}}$ & $\textrm{br@10}_{\textrm{novel}}$  \\ 
\hline
\multirow{3}{*}{n=5}  & FT-CE                                                 & 65.44 +/- 0.65                                                         & 74.57 +/- 0.16                                                       & 38.0 +/- 2.14~                                                          & 73.08 +/- 0.54                                                              & 82.33 +/- 0.24                                                        & 45.33 +/- 1.47                                                            \\
                      & PN                                                    & 66.47 +/- 0.58                                                           & 67.55 +/- 0.15                                                         & 63.24 +/- 2.34~                                                         & 75.28 +/- 0.54                                                                & 74.94 +/- 0.13                                                          & 76.28 +/- 2.13                                                            \\
                      & PCN (ours)                                            & 70.66 +/- 0.64                                                           & 74.47 +/- 0.18                                                         & 59.24 +/- 2.57                                                          & 79.10 +/- 1.04                                                                & ~81.41 +/- 0.29                                                         & 72.16 +/- 3.59                                                            \\ 
\hline
\multirow{3}{*}{n=10} & \multicolumn{1}{c}{FT-CE}                             & 69.86 +/- 0.46                                                           & 73.81 +/- 0.6                                                          & 58.0 +/- 3.22~                                                          & 77.9 +/- 0.60                                                                 & 81.91 +/- 0.18                                                          & 65.87 +/- 2.62                                                            \\
                      & \multicolumn{1}{c}{PN}                                & 67.51 +/- 0.39                                                           & 65.96 +/- 0.28                                                         & 72.16 +/- 1.66                                                          & 75.87 +/- 0.57                                                                & 72.94 +/- 0.27                                                          & 84.68 +/- 2.22                                                            \\
                      & \multicolumn{1}{c}{PCN (ours)} & 71.41 +/- 0.66                                                           & 70.37 +/- 0.16                                                         & 74.52 +/- 2.62                                                          & 79.93 +/- 0.47                                                                & 78.23 +/- 0.21                                                          & 85.04 +/- 2.00                                                            \\
\bottomrule
\end{tabular}
\label{tab:bal_main}
\end{table*}


\section{Per-class Accuracy}
\par\noindent
In Figure~\ref{fig:scatter} we provide a class-wise performance comparison between the PCN and $FT_{200}$-CE models as a scatter plot, in order to demonstrate their efficacies (shown here for the best performing PCN model evaluated with a train shot of 10 with $\textrm{mca}_{\textrm{base+novel}} = 50.92$, details in Table 1).

\par\noindent
We make the following observations. Overall metrics indicate that  FT-CE demonstrates slightly stronger average performance on base classes. For a large fraction of the base classes, both methods have similar performance. For the ones in which PCN performance is lower, there is usually a reasonable lower bound on the classification accuracy. In contrast, for novel classes, PCN performs better on average. Importantly, when FT-CE performance is lower than PCN, it is usually significantly lower. As an example is the novel class 'distal splitting hang nail' for which PCN performs significantly better.


\section{Additional Metrics}
\par \noindent
In table~\ref{tab:main} we provide recall@5 and recall@10 metrics for PN, PCN, and FT-CE approaches, for train shot $n=5$ and $n=10$. PCN performs on par with the FT-CE baseline and outperforms PN on these metrics. We note that since our test set is imbalanced, recall@k metrics unfairly reward strong performance on the head classes (which is observed with FT-CE). However, it is clear that PCN and PN models dominate in recall@k metrics on novel classes. \\
\par \noindent
To provide a fairer comparison, in Table~\ref{tab:bal_main} we report \textit{balanced} (or macro) recall@k metrics, wherein we compute recall@k for each class and average, treating each class as equally important. Here we clearly find PCN to outperform all baselines owing to strong performance across the board on base and novel classes.
